\documentclass[letterpaper, 10pt, conference]{ieeeconf}  % Comment this line out if you need a4paper

\IEEEoverridecommandlockouts                              % This command is only needed if 
\usepackage{multirow}
\usepackage{amsmath, amsfonts}
\usepackage{xcolor}
\usepackage{subcaption}
\usepackage[utf8]{inputenc}
\usepackage{tikz, tabularx}
\usepackage{overpic}
\usepackage{lipsum}
\usepackage{placeins}
\usepackage{graphicx}
\usepackage{booktabs}
\usepackage{censor}
\usepackage{caption}

\title{\LARGE \bf
Event-Driven Proactive Assistive Manipulation with Grounded Vision-Language Planning
}
\author{Fengkai Liu$^{1}$, Hao Su$^{1}$, Haozhuang Chi$^{2}$, Rui Geng$^{1}$, Congzhi Ren$^{1}$, Xuqing Liu$^{1}$, Yucheng Xu$^{1}$, \\ Yuichi Ohsita$^{1}$ and Liyun Zhang$^{3 \dagger }$% <-this % stops a space
\\[0.4em]
{\small $^{1}$The University of Osaka \quad $^{2}$Nanyang Technological University} \small \quad $^{3}$The University of Tokyo\\[0.2em]
{\small Email: \texttt{f-liu@ist.osaka-u.ac.jp}}
\thanks{$^{\dagger}$Corresponding author.  }%
\thanks{*This work was supported by “Program for Leading Graduate Schools” of The University of Osaka, Japan and by JST SPRING, Grant Number JPMJSP2138. }% <-this % stops a space
}

\begin{document}

\maketitle
\thispagestyle{empty}
\pagestyle{empty}

%%%%%%%%%%%%%%%%%%%%%%%%%%%%%%%%%%%%%%%%%%%%%%%%%%%%%%%%%%%%%%%%%%%%%%%%%%%%%%%%
\begin{abstract}
Assistance in collaborative manipulation is often initiated by user instructions, making high-level reasoning request-driven.
In fluent human teamwork, however, partners often infer the next helpful step from the observed outcome of an action rather than waiting for instructions.
Motivated by this, we introduce a shift from request-driven assistance to event-driven proactive assistance, where robot actions are initiated by workspace state transitions induced by human--object interactions rather than user-provided task instructions.
To this end, we propose an event-driven framework that tracks interaction progress with an event monitor and, upon event completion, extracts stabilized pre/post snapshots that characterize the resulting state transition. 
Given the stabilized snapshots, the planner analyzes the implied state transition to infer a task-level goal and decide whether to intervene; if so, it generates a sequence of assistive actions.
To make outputs executable and verifiable, we restrict actions to a set of action primitives and reference objects via integer IDs.
We evaluate the framework on a real tabletop number-block collaboration task, demonstrating that explicit pre/post state-change evidence improves proactive completion on solvable scenes and appropriate waiting on unsolvable ones. 
\end{abstract}

%%%%%%%%%%%%%%%%%%%%%%%%%%%%%%%%%%%%%%%%%%%%%%%%%%%%%%%%%%%%%%%%%%%%%%%%%%%%%%%%
% 问题
% 1. 描述过于泛，不够具体
% implicit intent: 描述不够准确；实际上人说明了他的需求，类似一种 text-driven，出现了dialogue，人类是主动的
% a) 基于明确需求；b) 基于隐含性需求；
% c) 人没有提示需求
% 2. 第一段，有些啰嗦；第一句没有价值；natural collaboration 过于抽象
% 例：具身智能经过了语言的..., 得到了很大的进步。但仍需要机器人主动地进行帮助。
% 考虑主要使用之前调研的文献，加一两个 a) 和 b) 的工作。
% 3. 创新点
% event-driven 和 command-free 两个连词不太学术。建议：a novel paradigm shift from the communication-driven(?) to event-driven。强调是范式或者框架。command-free 应该是具象而非概括的一部分。
% a novel framework of the event-based ~
% 创新点3，我们验证了我们的框架的有效性/trick
% 或者，affordable ID-grounded constrain，增强模型的 affordable, improved affordability ，（可实现性，运行更加可靠，合乎常理）
% 4. we validate the ~ 放在创新点之前
% 5. motivation 图
% a) ____ (代表工作名)
% 考虑pipeline? 图片清晰一些
% 摆盘有点问题
% 去掉机械臂底座，考虑俯视角
% 明确能一眼看出来是在做算数
% 人 ===> 机器人
% 人 =x=> 机器人
% b) ____ (代表工作名)
% c) ____ (proposed)
% f-liu_IROS2026

\section{Introduction}
\label{sec_intro}
Recent advances in embodied intelligence have enhanced robots' ability to interpret task instructions, ground them in perception, and generate executable actions in diverse manipulation settings~\cite{kim2024openvla, black2024pi0}. 
However, instruction following mainly addresses what to do once a specification is provided.
To move toward fluent teamwork, a robot should be able to decide when to intervene and what assistance to provide by inferring human intention, the unobserved internal states or goal-oriented constructs that underlie observable human behavior~\cite{hoffman2024inferring}. 
Recent reviews further suggest that proactive interaction can be facilitated by anticipation from observations and contextual information, often without explicit input from a human~\cite{den2024what}. 

\begin{figure}[t!]
    \centering
    \includegraphics[width=\linewidth]{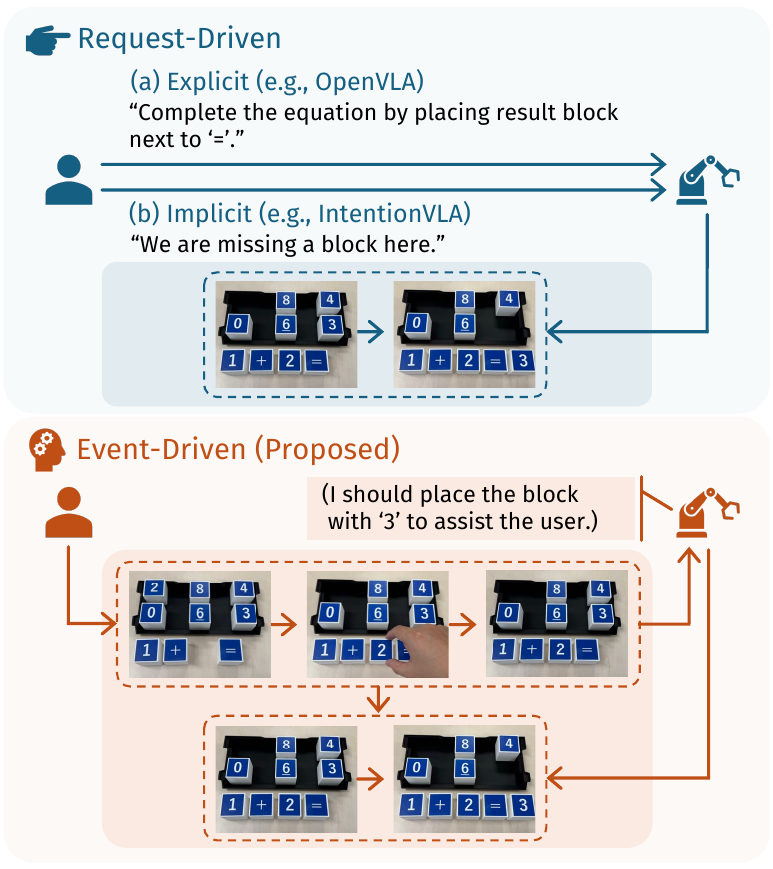}
    \caption{Motivation: request-driven vs.\ event-driven assistance. Request-driven systems initiate planning from requests provided by the user~\cite{kim2024openvla, chen2025intention}. Our event-driven setting instead triggers on human--object state transitions to infer the user’s goal and act without an additional user request.}
    \label{fig_motivation}
    \vspace{-1.5em}
\end{figure}

Translating this vision into practice reveals a bottleneck: many assistive systems remain request-driven, meaning that robot actions are initiated only after the human issues a request to the robot. 
Requests can be explicit, for instance a natural language command that specifies what to do~\cite{mahadevan2024generative,dai2024think}, or implicit, conveyed through underspecified verbal instructions~\cite{chen2025intention} or instructive nonverbal cues~\cite{tay2026intent}.
Even with large language models (LLMs) and vision-language models (VLMs), whose strong contextual reasoning has been used to support goal inference~\cite{zhang2023large}, prior assistive systems commonly invoke these models only after receiving such requests, preserving a request-driven interaction pattern~\cite{chen2025intention}. 
Consequently, interaction often remains in a request--response loop, which limits proactive assistance when step-by-step guidance is absent. 

Motivated by this gap, we move from request-driven assistance to an event-driven proactive assistance setting. % this paper advocates shifting the paradigm from request-driven to event-driven proactive assistance. 
In human-human teamwork, partners often do not wait for explicit instructions after every step. Instead, they use the observed outcome of an action to infer the likely next sub-goal and decide whether to assist. 
Motivated by this interaction pattern, we define an event as the human--object interaction interval initiated by the human that induces a sustained workspace state change. 
Its completion serves as a natural trigger for proactive robotic assistance.
This design choice is motivated by recent results showing that VLMs can interpret dynamic environmental changes and use them to guide adaptive planning~\cite{cai2025vision}. 
Under this framework, the robot monitors the workspace for human--object interaction events and captures stabilized pre/post observations around each event. 
At event completion, it queries a VLM with the stabilized pre/post evidence to infer a task-level goal and generate the corresponding assistive action(s), without requiring an additional user request specifying the next step.
To improve safety and mitigate errors from generative planning, we incorporate multi-layer safeguards in the control loop, including schema-constrained outputs, ID-grounded references, and local verification.
We validate the proposed framework on a real tabletop manipulation platform in scenarios without step-by-step user instructions, and report quantitative results on success rates and failure types on a shared evaluation set.
Our contributions are threefold:
\begin{itemize}
    \item \textbf{A paradigm shift from request-driven assistance to event-driven proactive assistance.}
    % \item \textbf{A formulation of proactive assistance from request-driven to event-driven.}
    We cast proactive assistance in our setting as an event-driven process in which robot actions are initiated by workspace state transitions caused by human--object interactions, rather than user-provided requests or task specifications. 

    \item \textbf{An event-driven embodied framework for goal inference and assistive action synthesis.}
    We propose an event-driven embodied framework that uses an event state machine to detect interaction events, select compact pre/post snapshots, and invoke VLM reasoning at event-triggered decision points for task-level goal inference and assistive action generation. 

    \item \textbf{An ID-grounded executable interface for reliable and affordable deployment.}
    We introduce an ID-grounded executable interface that grounds VLM outputs in structured scene references and constrained robot actions, reduces execution ambiguity, and supports low-overhead verification, including object-ID existence checks and post-execution outcome verification, thereby improving system reliability and affordability. 
\end{itemize}
 
\section{Related Work}
\label{sec_related_work}

Building on Sec.~\ref{sec_intro}, we organize related work by what initiates robot actions.
We group prior systems into request-driven assistance and event-driven assistance.
We first review request-driven systems, distinguishing explicit requests from requests embedded in indirect cues, and then discuss event-driven systems that leverage interaction outcomes.

\subsection{Request-Driven Assistance}
\label{sec_related_request_driven}

A large body of collaborative robot systems remains request-driven: even when strong foundation models enhance the perception of human behavior and contextual reasoning, robot actions are still typically initiated only after user requests become available.

\subsubsection{Explicit Request}
Systems based on explicit requests initiate robot actions from user-provided commands, requests, or task specifications.
In such settings, language-to-plan pipelines condition action generation on the given instruction and ground candidate steps for execution~\cite{ahn2022saycan}. 
Other systems compile instructions into executable programs that call perception and state-estimation modules and reusable skills~\cite{liang2023code, zhang2023panoptic, mahadevan2024generative, zhang2024panoptic, hoffmeister2025towards}, and improve robustness through explicit feedback and re-planning loops~\cite{dai2024think}.

Many recent embodied foundation-model systems follow the same interaction pattern.
RT-2 scales language- and vision-conditioned manipulation and improves generalization through large-scale data and pretraining transfer~\cite{zitkovich2023rt2}.
PaLM-E integrates language, perception, and robot state for embodied reasoning~\cite{driess2023palme}, while grounding interfaces connect model outputs to 3D workspaces for execution~\cite{huang2023voxposer}.
Recent work also targets ambiguous instructions with more structured spatial reasoning~\cite{wan2025infer, huang2025graphcotvla}, and generalist robot models continue to expand capability coverage~\cite{geminirobotics2025}.
Despite differences in architecture and execution interface, these systems typically begin high-level reasoning only after an instruction or a stated goal.

\subsubsection{Implicit Request}
Another request-driven line of work initiates robot actions from indirect cues that embed implicit requests: without an explicit task command, the robot receives such cues and infers the implicit request they carry to initiate actions.
Typical cues include ambiguous expressions~\cite{chen2025intention}, gaze~\cite{tay2026intent}, gestures~\cite{lai2025zeroshot}, or facial expression and co-speech motion cues~\cite{sha20253dfacepolicy, sha20263dgespolicy}.
Classic approaches perform online goal or intent inference and adapt robot behavior for fluency and efficiency, often using probabilistic estimation or prediction over a fixed goal set~\cite{liu2016data,hoffman2024inferring,liu24anticipatory,ghose2025adapting}.
Related ideas also appear in shared autonomy, where the robot maintains a belief over user goals and assists under uncertainty~\cite{brooks2020visualization, jonnavittula2022communicating}.
Activity and action prediction methods further support proactive preparation by anticipating near-future human behavior from motion and object context~\cite{mascaro2023hoi4abot, lazzari2025pace}.

These methods are effective in constrained settings, but many rely on predefined goal spaces or phase structures, partly because annotations can be subjective and incomplete, which limits open-ended assistive manipulation~\cite{hoffman2024inferring, zhang2026simlabel}.
Recent work explores foundation-model semantics to reduce reliance on fixed labels.
LLMs have been used as zero-shot models of human behavior for planning~\cite{zhang2023large}, and VLA/VLM-based systems have been developed for intention reasoning and human state monitoring in complex environments~\cite{zhang2024microemo, chen2025intention, wang2025roboomni, zhang2025mindpower, chi2025vlmdm, lai2025zeroshot, tay2026intent, zhang2026acoustemo}.
In practice, many such systems still rely on communication or coarse task framing to initiate or stabilize inference~\cite{chen2025intention, wang2025roboomni, lai2025zeroshot}.
Therefore, even when the request is indirect and under-specified, robot actions are still commonly initiated by requests rather than by evidence from observed state transitions. 
Moreover, the semantics of indirect cues can be subjective, and human judgments about the underlying request may vary across observers~\cite{zhang2025qumatl, zhang2026unified, zhang2026qumab}. 

\subsection{Event-Driven Assistance}
\label{sec_related_event_driven}
A related line of work leverages workspace changes at decision points to update robot actions.
Cai et al. provide an early multi-VLM collaboration system that infers human intent and updates robot actions from changing workspace observations, including a core task description as coarse task/scene framing, within an alternating human--robot interaction protocol that supplies step-level decision points~\cite{cai2025vision}.
These choices suggest a broader design space for event-driven assistance; our work advances it by studying initiating actions at inference time without user task specifications and by implementing runtime event boundary detection with compact pre/post evidence selection for VLM reasoning.

In our setting, robot actions at inference time are not initiated by user-provided task specifications, nor by cues that imply a request.
Instead, actions are initiated by human--object interaction outcomes that produce observable workspace state changes, and the post-event configuration serves as the primary evidence for selecting the next assistive action.
This positioning motivates the event-centric pre/post reasoning framework for proactive assistive manipulation developed in Sec.~\ref{sec_method}. 

% overview 的 execution 图考虑放实物图，event 前后，机械臂执行结束后使用真实图，作为一个模块
% Sensing Result 到 Geo-Grounding 增加连接
% Geo-Grounding 画得具体一些
% Policy Generation 画得小一些
% 文字减少一些
% Event State Machine 太小？
% 不能叫 annotation，考虑去掉？降低文章价值
% 考虑 $I_{ann}^{+}$(FastSAM)或类似表述
% frozen 表示的是有一部分参与训练的，目前使用的都是 pre-trained 的模型，不合适
% 考虑 总图-分图 的模式

\section{Methodology}
\label{sec_method}
This section presents an event-driven framework for proactive assistance that couples event-level VLM planning with on-device grounding. 
We first define the problem setting and its key challenges, then introduce the system architecture. 
We then detail the local pre-processing, cloud planning, and local post-processing stages of the framework. 

\subsection{Problem Setting: Event-Driven Proactive Assistance}
\label{sec_method_formulation}
We focus on proactive assistive manipulation in human--robot shared workspaces.
An interaction event is a human--object interaction interval that induces a sustained workspace state change, with onset and offset times detectable from activity cues. 
We assume that the outcome of such interactions provides task-relevant evidence.
Accordingly, we represent the human intention relevant for assistance as a task-level goal for assistance, namely a latent goal configuration consistent with the observed pre/post state transition, e.g., desired object placements and qualitative spatial relations. 
Under this view, the system monitors interaction events and, once an event terminates, infers an assistance goal from the state transition and decides whether to act or wait. 
Triggering planning at event completion provides a more informative and less ambiguous observation than intermediate frames during ongoing interaction. 
This yields a setting where assistance initiation does not rely on user requests for the current step: at each decision point, the robot initiates assistance based on the observed state transition. 

\subsection{System Architecture}
\label{sec_method_architecture}
\begin{figure*}[t]
    \centering
    \includegraphics[width=\linewidth]{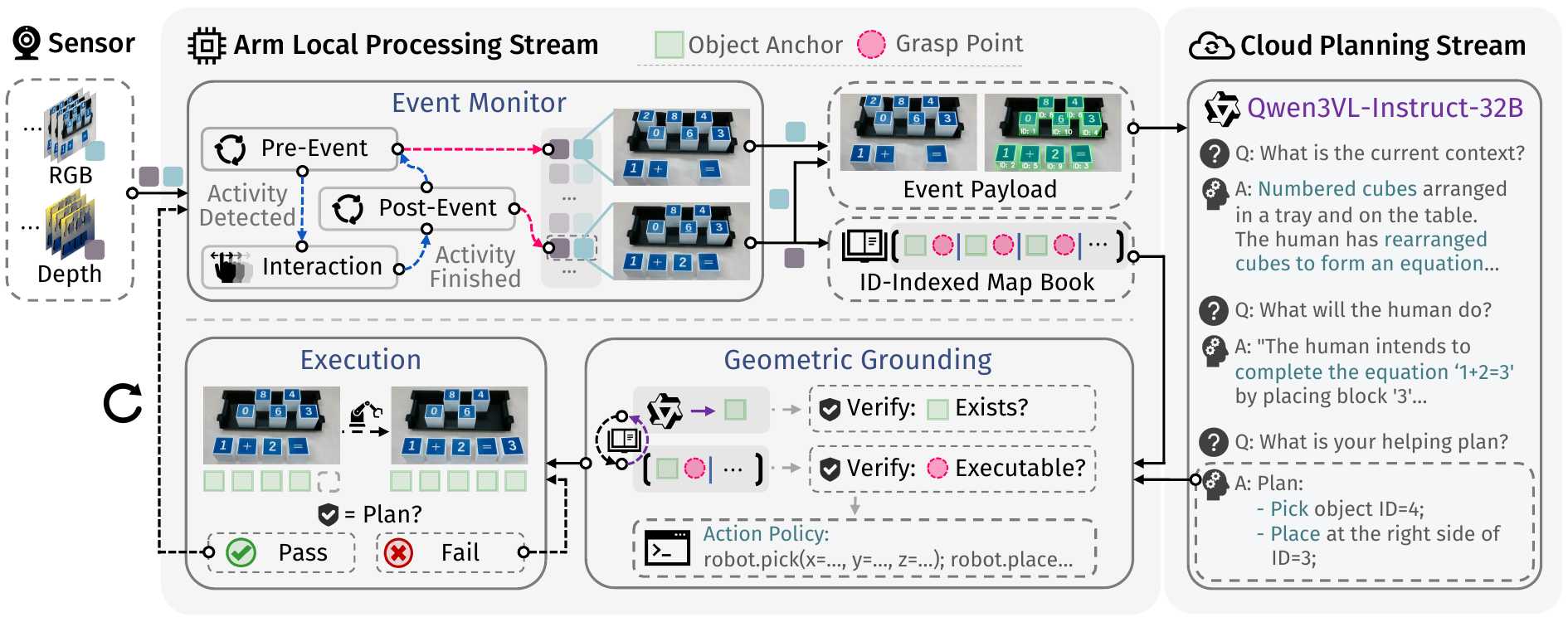}
    \caption{System overview. The arm-side local stream monitors events, constructs an event payload, and executes grounded actions with verification; a cloud VLM performs event-level planning and returns an ID-indexed symbolic plan. }
    \vspace{-1.5em}
    \label{fig_overview}
\end{figure*}

A key challenge in realizing event-driven assistance is semantic generalization to long-tailed event cues. 
The task cues revealed by an interaction event are diverse and long-tailed, and are difficult to cover with a fixed set of hand-crafted rules. 
To handle this variability, we use a VLM as an event-level planner, allowing the system to infer a plausible assistance goal from pre/post observations and propose corresponding actions. 

Using a VLM for planning, however, introduces an execution gap: model outputs are often open-ended and may be inconsistent, spurious, or not directly actionable for control.
We address this by organizing the planner input as a compact, structured event payload and restricting the planner output to a structured action list over a small set of allowed primitives. 
Using ID-indexed symbolic references further reduces the impact of such spurious outputs by avoiding direct metric coordinate prediction from the VLM. 
Before execution, the local controller validates the structured output with schema checks and runtime guards, and rejects plans that are invalid or non-executable.

Figure~\ref{fig_overview} illustrates the overall architecture, which separates the system into an arm-side local processing stream and a cloud planning stream.
The local stream runs continuously for sensing, while the cloud stream is invoked only when a human-initiated interaction event completes, resulting in alternating human action phases and robot response phases. 

On the local stream, the system monitors the RGB--D video and maintains an event state machine. 
At event onset, the system selects the most recent stable observation as the pre-event snapshot $I^{-}$; after the event terminates, it records a stable post-event snapshot $I^{+}$.
The post-event snapshot is then parsed locally to detect object instances, assign integer IDs, and construct a post-event object map $\mathcal{T}^{+}$ that stores the geometric information required for subsequent grounding and manipulation.
The local stream sends the event snapshot $(I^-, I^+)$, where $I^+$ includes ID-indexed object references, to the cloud planner as a payload. 

On the cloud stream, given $(I^{-}, I^{+})$, the cloud VLM returns a structured action list over a set of primitives.
Actions are expressed using object IDs only: grasping references a target ID, and placement is specified relative to a reference ID with a qualitative spatial relation.

Back on the local stream, the local controller validates the received plan, resolves all referenced IDs using $\mathcal{T}^{+}$, grounds symbolic actions into metric targets, and executes them on the robot.
Execution outcomes are checked from refreshed observations; failures trigger re-planning and, if necessary, a recovery procedure before returning to the event-driven monitoring loop.

In the following subsections, we describe these three stages of the framework in detail.

\subsection{Local Pre-Processing Stream}
\label{sec_method_local_pre}
We describe how the local side detects interaction events from a continuous RGB--D stream, selects stable pre/post snapshots for the payload, and constructs the post-event object map $\mathcal{T}^{+}$ used to ground the ID-indexed plan returned by the cloud planner. 

\subsubsection{Event Monitoring and Snapshot Selection}
\label{sec_method_local_event}
% Let $\{(I_t^{rgb}, I_t^{d})\}_{t\ge 0}$ denote the incoming RGB--D stream.
Event progress is tracked by a three-phase state machine:
pre-event $\rightarrow$ human action $\rightarrow$ post-event.
Over the manipulation workspace, an activity signal $\rho(t)$ is computed from activity cues, specifically the optical-flow magnitude in our implementation. 
An event onset is declared when $\rho(t)$ persistently exceeds an onset threshold; the event offset is declared when $\rho(t)$ drops and remains below an offset threshold under a stability constraint. 

Stable observations around the detected event onset and offset are obtained using two bounded snapshot buffers, one for pre-event frames and another for post-event frames. 
At onset, the most recent stable frame in the pre-event buffer is selected as the pre-event snapshot $I^{-}$. 
After the event terminates and the scene becomes stable, the first stable frame is written to the post-event buffer and used as the post-event snapshot $I^{+}$. 
This buffering selects stabilized keyframes around event boundaries and avoids transmitting frames with ongoing manipulation. 
We therefore transmit only $(I^{-}, I^{+})$ as compact evidence for planning.

\subsubsection{Post-Event Parsing and Local Object Map Construction}
\label{sec_method_local_map}
After an event terminates, the local stream parses the post-event snapshot $I^{+}$ to obtain object instances for manipulation and grounding. 
Parsing the post-event scene ties symbolic references to the immediately executable workspace state and avoids unnecessary computation on the pre-event snapshot. 
This step can be instantiated with any off-the-shelf instance segmentation model; we use FastSAM to extract instance masks and bounding boxes from $I^{+}$ for its efficiency in on-device deployment. 
The parsing output yields a set of object instances
\begin{equation}
\mathcal{D}(I^{+}) = \{(b_k, m_k, s_k)\}_{k=1}^{K},
\end{equation}
where $b_k$ is a 2D bounding box, $m_k$ is an instance mask, and $s_k$ is a confidence score. 
Each detected instance is assigned an integer ID for symbolic reference in the current event, yielding an ID-indexed object map 
\begin{equation}
\mathcal{T}^{+}[\textit{id}] = (b, m, \mathbf{a}, \mathbf{g}),
\end{equation}
where $\mathbf{a}$ is a 2D anchor for referencing and $\mathbf{g}$ is a grasp proposal computed from RGB--D geometry within the instance.
The map $\mathcal{T}^{+}$ remains local and is used to resolve the IDs returned by the cloud planner and to support geometric grounding. 

With the stabilized snapshot pair available after an event, the local stream forms the event payload
\begin{equation}
E = (I^{-}, I^{+}),
\label{eq_event_payload}
\end{equation}
and transmits $E$ to the cloud planner once the post-event scene has stabilized.

% \subsubsection{Event Payload Generation}
% \label{sec_method_event_payload}
% With the snapshot pair $(I^{-}, I^{+})$ available after an interaction event, the local side forms an event payload for cloud planning.
% In our interface, the payload is the event summary
% \begin{equation}
% E = (I^{-}, I^{+}),
% \label{eq_event_payload}
% \end{equation}
% which is transmitted to the cloud VLM planner once the post-event scene has stabilized. 

\subsection{Event-driven VLM Planning}
Upon receiving the event payload $E$ from the arm-side platform, the cloud module invokes a VLM for event-level reasoning and planning.
Unlike request-driven settings, we do not provide the VLM with an explicit user request that specifies the next step.
Instead, we provide the event context via the pre/post snapshots in the event payload and constrain the planner with a fixed contract that specifies the primitive set and required structured output format. 
The cloud planner returns a structured action list over a restricted primitive set (\texttt{pick}, \texttt{place}, and \texttt{wait}), with all object references expressed using IDs.
Grasping is specified by a target object ID, while placement is specified by a reference object ID together with a qualitative relation indicating the desired placement relative to the reference.
Free-form text is treated as non-executable logging, and the controller consumes only the structured action list.

\subsection{Local Post-Processing}
After the cloud planner returns an ID-indexed symbolic action list, the arm-side system performs local post-processing to turn symbolic actions into executable robot motion, and then checks whether the intended outcome is achieved.

The local controller first checks the received plan against basic execution constraints: actions must belong to the allowed primitive set, all referenced IDs must exist in the current post-event object map, and each action must be feasible under the current scene geometry and workspace constraints.
After validation, each action is bound to entries in the post-event object map $\mathcal{T}^{+}$.
For each object ID, $\mathcal{T}^{+}$ provides a reference anchor for relational placement and a grasp proposal computed from post-event RGB--D geometry.
These quantities are converted into metric targets in the robot base frame via RGB--D back-projection and the calibrated camera-to-base transform.

For grasping, a \texttt{pick(target\_id)} action specifies only the target ID.
The local controller resolves the ID in $\mathcal{T}^{+}$, retrieves the stored grasp proposal, and executes the grasp with its on-device motion primitives.
For placement, a \texttt{place(reference\_id, relative\_pos, offset\_scale)} action is specified by the cloud planner using a reference object ID, a qualitative relation, and a scalar offset parameter.
The local controller resolves reference ID to obtain the reference anchor, maps the qualitative relation to a planar direction around the anchor, and converts offset scale into a metric displacement using locally available post-event geometry, yielding a placement target.
In this way, the cloud output remains ID-indexed and relational, while all coordinate resolution is performed on the local side.

After execution, the system captures refreshed observations, updates the local object map, and performs outcome verification. 
Verification checks whether the qualitative relations implied by the symbolic plan hold in the updated scene. 
A failure is declared when expected relations are violated or referenced objects cannot be recovered in the refreshed observations. 
If verification succeeds, the system returns to event monitoring. 
If verification fails, the system retries execution. 
We allow a bounded number of retries; persistent failures trigger a recovery procedure before returning to event monitoring. 

% 参考别人的 metric 修改目前的
% 和 baseline 的比较
% command-driven (openvla-style) ，只给 command，如把 3 放到正确的位置
% implicit?? ，给 prompts 如 do math
\section{Experiments}
\label{sec_evaluation}
In this section, we evaluate the effectiveness of our event-driven proactive assistance framework in a real tabletop collaboration task.
We assess whether, without user-provided requests, our framework can leverage event-aligned pre/post observations to make appropriate assist decisions and generate the next assistive action that is correct under our task rules.
Video demonstration is provided in the \textbf{supplementary material}. 

\begin{figure*}[t!]
    \centering
    \includegraphics[width=\linewidth]{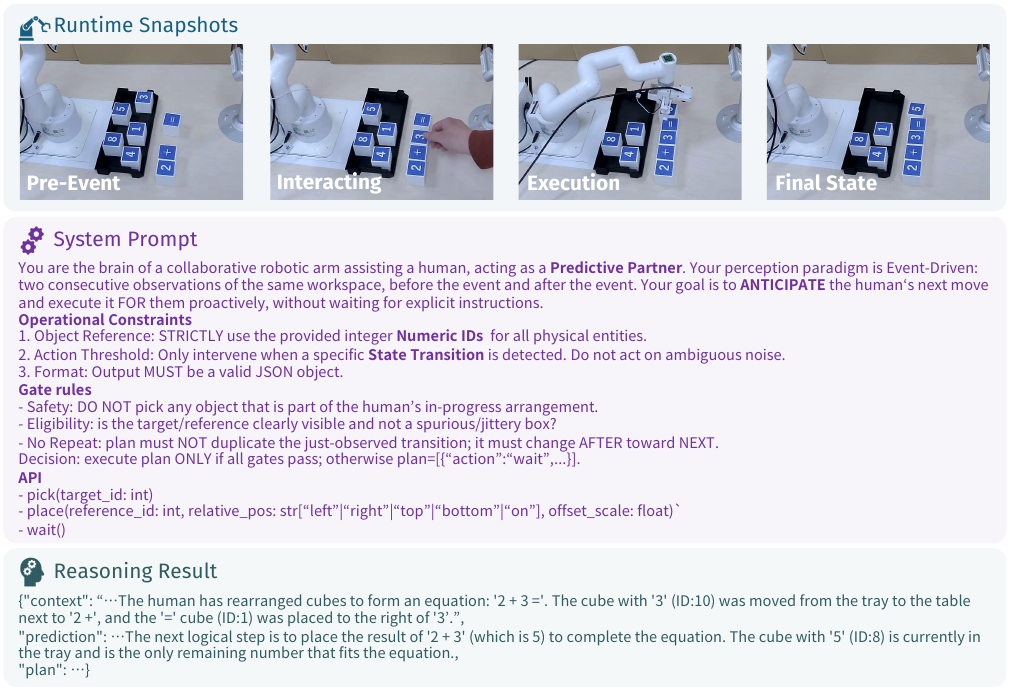}
    \caption{Qualitative example of event-driven proactive completion in the tabletop number-block task. The top row shows runtime snapshots (pre-event, human interaction, robot execution, and the final state). The middle row shows the system prompt. The bottom row shows the planner's event interpretation and the resulting structured plan, completing $2{+}3{=}5$ by placing ``5'' to the right of ``=''.}
    \label{fig_case_study}
    \vspace{-1.5em}
\end{figure*}

\subsection{Experimental Setting}
\subsubsection{Task Definition}
\label{sec_eval_task}
We consider a tabletop number-block collaboration task as our evaluation case study. 
In this task, a human builds an arithmetic expression step by step by placing operands and operators such as ``+'' and ``=''. 
Each block corresponds to a single digit or operator symbol, and our evaluated scenes cover multiple arithmetic patterns, including addition, subtraction, multiplication, and division. 
The robot receives no additional user request specifying the next step; it decides when and how to assist based on observed workspace changes. 
We evaluate two case types. 
\textbf{Solvable} scenes contain the correct result block among the remaining candidates, in which case the robot should proactively place it to complete the expression, typically to the right of ``=''.
\textbf{Unsolvable} scenes do not contain a valid completion under the current candidates, in which case the robot should output wait until a new observable change occurs.

\subsubsection{Baselines}
Under the same task rules in Sec.~\ref{sec_eval_task}, we compare our event-driven framework with  controlled variants that isolate the effects of observation content and triggering schedule. 
\textbf{Post-only} removes the pre-event evidence and uses only the post-event snapshot at the same detected event boundary, testing whether a single stabilized post-state is sufficient for assist decisions.
\textbf{Always-on} removes event-trigger gating and queries the planner periodically on the video stream; in our implementation, we provide a short temporal context by concatenating five sampled frames (one every five frames from 30\,FPS, i.e., a 0.83\,s window) for each query. 
Each trial is capped at three queries; otherwise it is labeled as failure. 
\textbf{Request-driven} follows the paradigm discussed earlier: given a user instruction, it conditions planning on a single observation of the current workspace state.
For controlled comparison, we use the stabilized post-event snapshot as the workspace observation for each trial.

The system prompt differs in a short block that specifies the observation format and the corresponding paradigm statement: proposed uses ``two consecutive observations of the workspace before and after an interaction'' (refer to Fig.~\ref{fig_case_study}), whereas post-only and request-driven use ``one observation of the current workspace state'' and remove the event-driven description; always-on uses ``a sequence of observations of the workspace''. 
Request-driven additionally provides task specification via a one-sentence user instruction (``Complete the equation by placing the appropriate block.''). 
% We additionally examine Direct-coordinate as an interface contrast, which removes ID-based grounding and asks the VLM to output action coordinates directly.
% We exclude it from the main quantitative table because it does not share the same action contract and verification path as the ID-and-relation interface, and its zero-shot coordinate outputs are often unstable; we summarize typical failure modes to motivate our interface choice  in Sec.~\ref{sec_eval_qual}.
% Unless explicitly stated, all compared methods use the same VLM backbone, prompt template, detector/tracker outputs, action parser, robot execution stack, and verification pipeline as the full system.

\subsubsection{Metrics}
We evaluate collaborative behavior using outcome verification on the final workspace state.
Each trial corresponds to one decision point and is labeled by the resulting behavior under the task rules.
A trial is counted as successful if the robot performs the correct behavior for the current case type.
In solvable scenes, success requires executing an assistive action that leads to a verified correct completion, such as selecting the correct result block and placing it in the intended relation to ``=''.
In unsolvable scenes, success requires waiting without manipulating any block, and explicitly indicating in the planner output that no valid completion exists under the current candidate set. 

To better localize errors, we additionally separate failures into reasoning failures and execution failures.
A reasoning failure occurs when the high-level decision, as specified by the planner output, is incorrect under the task rules, including an incorrect wait/act decision, choosing an incorrect result block, or specifying an incorrect target relation. 
An execution failure occurs when the high-level decision is correct but the physical execution does not achieve the verified relation in the final state.
We report success rates (successful trials / total trials) and the corresponding failure breakdown for both solvable and unsolvable subsets.

\subsubsection{Experimental Platform}
We evaluate the full system on a Yahboom JetCobot platform (Jetson Orin 8GB, Elephant MyCobot 280 for Arduino) with an external RGB-D camera (RealSense D435i) observing the tabletop workspace.
VLM reasoning is performed on a PC equipped with an RTX 6000 Ada GPU, while event pre-processing and post-processing run on the robot side.

\subsection{Qualitative Example: Completion of $2{+}3{=}5$}
\label{sec_case_study}

Fig.~\ref{fig_case_study} illustrates an example where the system completes a partially formed expression after a human action. 
The human rearranges number blocks to construct ``$2{+}3{=}$'', moving the ``3'' block (ID:10) from the tray to the workspace and placing the ``='' block (ID:1) to the right of ``3''. 
Once the workspace stabilizes, the robot-side event module forms a pre/post snapshot pair around this detected state transition (top row in Fig.~\ref{fig_case_study}) and invokes the cloud planner with the tracked object IDs. 

Given the pre/post evidence, the planner infers that the next helpful step is to complete the equation by placing the result ``5'' to the right of ``=''. 
It selects ``5'' from the remaining candidates and outputs a structured plan that picks the ``5'' block and places it to the right of ``=''. 
The robot executes the plan, and the final state is checked by the outcome verification module to confirm that the intended relation is achieved and the expression is correctly completed.

\subsection{Quantitative Results}
\label{sec_eval_quant}
\subsubsection{Overall Outcomes}
\begin{figure}[t]
    \centering
    \includegraphics[width=\linewidth]{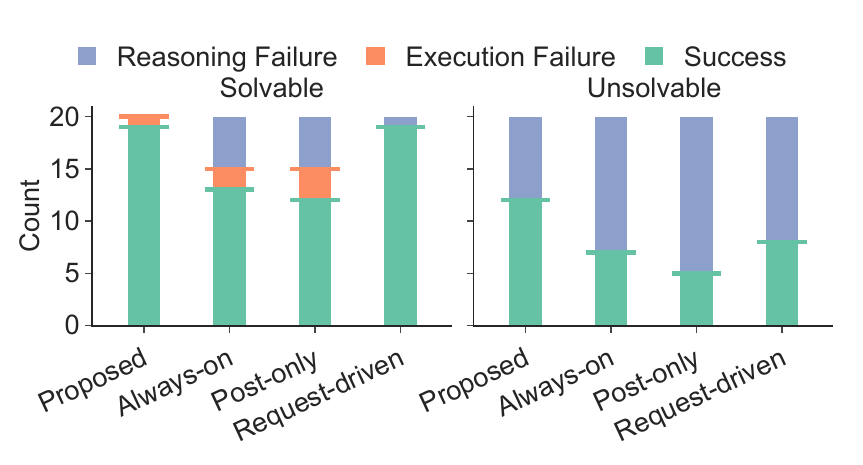}
    \caption{Quantitative results: outcome proportions. Each bar shows the fraction of successes and failures.}
    % , with failures decomposed into execution failures and reasoning failures
    \label{fig_proportions}
    
    \vspace{3pt}
    \centering
    \normalsize
    \definecolor{SetTwoSuccess}{RGB}{102,194,165} % Set2[0]  (Success)
    \definecolor{SetTwoRSR}{RGB}{252,141, 98}     % Set2[1]  (used for RSR anchor)
    \captionof{table}{Quantitative results: execution success rate (ESR) and reasoning success rate (RSR), reported in 0--1. }
    \vspace{2pt}
    \begin{tabular}{lcccc}
        \toprule
        \multirow{2}{*}{Method} &
        \multicolumn{2}{c}{Solvable} &
        \multicolumn{2}{c}{Unsolvable} \\
        \cmidrule(lr){2-3}\cmidrule(lr){4-5}
        & \textcolor{SetTwoSuccess}{\textbf{ESR}}$\uparrow$
        & \textcolor{SetTwoRSR}{\textbf{RSR}}$\uparrow$
        & \textcolor{SetTwoSuccess}{\textbf{ESR}}$\uparrow$
        & \textcolor{SetTwoRSR}{\textbf{RSR}}$\uparrow$ \\
        \midrule
        Proposed   & \textbf{0.95} & \textbf{1.00} & \textbf{0.60} & \textbf{0.60} \\
        Always-on  & 0.65 & 0.75 & 0.35 & 0.35 \\
        Post-only  & 0.60 & 0.75 & 0.25 & 0.25 \\
        Request-driven   & \textbf{0.95} & 0.95 & 0.40 & 0.40 \\
        \bottomrule
    \end{tabular}
    \label{tab_success_rate}
    \vspace{-1.5em}
\end{figure}

Fig.~\ref{fig_proportions} and Table~\ref{tab_success_rate} summarize outcomes over 40 trials (solvable 20 trials, unsolvable 20 trials).
Table~\ref{tab_success_rate} reports ESR and RSR (defined in the table caption) on the solvable and unsolvable subsets. 

On solvable scenes, proposed reaches RSR$=1.00$ in our evaluation set, indicating that event-aligned pre/post observations provide sufficient evidence for stable high-level decisions under the task rules.
The only error on this subset is an execution failure: although the reasoning is correct, the local arm execution causes an unintended collision with nearby blocks. 
By comparison, post-only and always-on exhibit lower reasoning and execution success on solvable scenes.
This gap can be attributed to reduced change evidence (single post-state) and unaligned querying that may observe transient or incomplete context, both of which make inferring the human's intended completion more difficult.
Request-driven achieves comparable performance on solvable scenes (ESR$=0.95$, RSR$=0.95$), suggesting that conditioning on a user instruction together with a single stabilized post-state can often resolve the next assistive step when a valid completion exists. 

On unsolvable scenes, success corresponds to abstaining with a no-solution indication; therefore ESR equals RSR on this subset.
Proposed improves the wait decision accuracy on this subset; the remaining failures exhibit consistent patterns, which we analyze next.

\subsubsection{Failure Mode Analysis}
\begin{figure}[t]
    \centering
    \includegraphics[width=\linewidth]{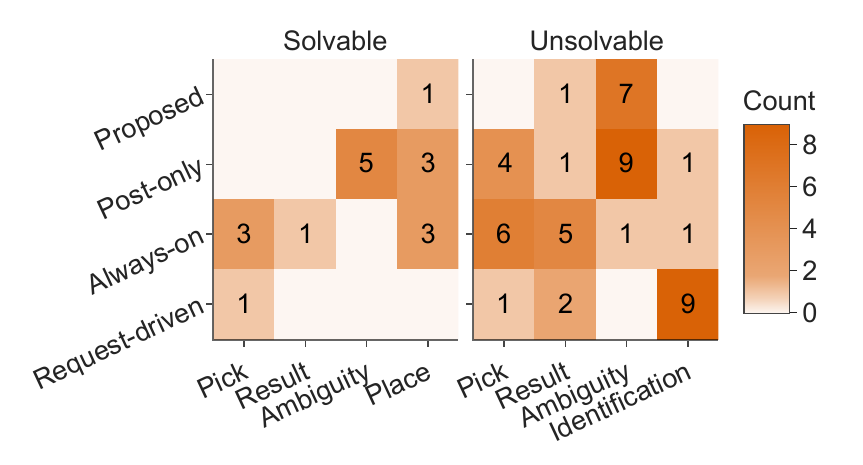}
    \caption{Failure taxonomy for solvable and unsolvable cases.
    \textbf{Pick}: unintended manipulation of a block that should remain fixed.
    \textbf{Result}: incorrect arithmetic completion.
    \textbf{Ambiguity}: insufficient evidence for a confident next assistive move.
    \textbf{Place}: final placement fails the desired spatial relation.
    \textbf{Identification}: perception-level misrecognition of symbols.}
    \label{fig_failure_taxonomy}

    \vspace{3pt}
    \centering
    \normalsize
    \captionof{table}{Planner call counts per successful trial. }
    \vspace{2pt}
    \begin{tabular}{lcc}
        \toprule
        Method & Solvable$\downarrow$ & Unsolvable$\downarrow$\\
        \midrule
        Proposed & $\textbf{1.00}$ & $\textbf{1.00}$ \\
        Always-on  & $1.77 \pm 0.53$ & $1.29 \pm 0.24$ \\
        \bottomrule
    \end{tabular}
    \label{tab_planner_calls}
    \vspace{-1.5em}
\end{figure}

Fig.~\ref{fig_failure_taxonomy} characterizes representative failure modes beyond the coarse success/failure outcomes.

On solvable scenes, failures of the controlled variants concentrate in a few interpretable categories.
Post-only failures are primarily due to ambiguity (Ambiguity:5, Place:3).
With only a single post-event snapshot, the observation often does not reveal which objects changed and for what purpose, making the human's intent under-specified; this uncertainty can lead to conservative waiting. 
Always-on exhibits a broader profile (Pick:3, Result:1, Place:3), which is largely explained by window alignment: when the sampled frame sequence captures a complete event, it can support correct inference similar to proposed, whereas sequences that overlap with the interaction frequently contain transient hand/object motion and intermediate object poses, leading to inaccurate action targets and relation specifications that manifest as unintended picks and failed placements. 
Request-driven shows an unintended manipulation (Pick:1), consistent with occasional inconsistency in goal-conditioned planning; this tendency becomes more pronounced on unsolvable scenes. 

On unsolvable scenes, proposed failures are dominated by Ambiguity (Ambiguity:7). 
In our observations, these cases typically arise when the evidence is insufficient to certify infeasibility, leading the planner to hypothesize an alternative intent such as workspace re-organization and propose moving recently placed blocks to `tidy'' the layout. 
A small number of failures also attempt an incorrect completion (Result:1), where a negative result is required but no negative sign candidate is available and the planner effectively ignores the missing sign. 
Post-only failures are primarily due to ambiguity (Ambiguity:9) with occasional unintended picks (Pick:4) and rare identification error (Result:1, Identification:1), consistent with the fact that a single post-state provides limited evidence to infer infeasibility and often under-specifies what has just changed. 
Always-on failures include both unintended picks and incorrect completions (Pick:6, Result:5), with additional errors from ambiguity and identification error (Ambiguity:1, Identification:1); as discussed above, these failures are largely attributable to window alignment that can include transient intermediate states rather than a clean event-level context. 
Request-driven failures are dominated by Identification (Identification:9) along with occasional incorrect completions (Result:2) and unintended picks (Pick:1). 
We observe reasoning inconsistency in the instruction conditioned planner when the current candidate set does not support a consistent completion. 
In such cases, the model may still commit to a completion attempt and output a plan that conflicts with the task logic instead of abstaining, making request-driven prompting less reliable on unsolvable scenes.

\subsubsection{Runtime Cost} 
Table~\ref{tab_planner_calls} reports planner-call counts per successful trial.
Proposed uses exactly one planner call per successful trial, and post-only and request-driven likewise operate in a single-query mode in our implementation.
Always-on incurs higher call counts because it queries the planner periodically until a sampled window yields a confident decision, increasing inference overhead even on successful trials. 

\subsubsection{Summary}
Overall, event-aligned pre/post observations yield more reliable reasoning in this task, while the remaining failures are dominated by placement robustness in solvable scenes and misinterpretations in unsolvable scenes.

\section{Conclusion}
\label{sec_conclusion}
We present an event-driven paradigm shift in proactive assistance for our setting, where robot action is initiated by workspace state transitions induced by human--object interactions rather than by user-provided requests.
Building on this paradigm, we develop an event-driven embodied framework that uses an event state machine to capture stabilized pre/post snapshots and invokes VLM reasoning at event-triggered decision points for goal inference and assistive action synthesis.
On a real tabletop number-block collaboration task, explicit pre/post evidence consistently outperforms baselines.
Future work will improve end-to-end robustness, extend anticipation to longer horizons for stronger proactivity, and validate the framework across broader scenes and tasks.

\bibliographystyle{IEEEtran}
\bibliography{references}

\end{document}